\documentclass[letterpaper]{article} %
\usepackage{sty/aaai2026}  %
\usepackage{times}  %
\usepackage{helvet}  %
\usepackage{courier}  %
\usepackage[hyphens]{url}  %
\usepackage{graphicx} %
\usepackage{natbib}  %
\usepackage{caption} %
\usepackage[T1]{fontenc}
\usepackage{amsmath,amssymb,amsthm}
\usepackage{algorithm}
\usepackage{booktabs}
\usepackage{multirow}
\usepackage{cleveref}
\usepackage{siunitx}
\usepackage{sty/custom}
\usepackage{algorithmicx}
\usepackage[noEnd=true,indLines=true,commentColor=black]{sty/algpseudocodex}
\usepackage[most]{tcolorbox}
\usepackage{soul}
\usepackage{mathtools}
\sethlcolor{cyan}
\newtheorem{definition}{Definition}

\usepackage{graphicx}   %
\usepackage{pgfplots}   %
\pgfplotsset{compat=newest}

\usepackage{newfloat}
\usepackage{listings}
\DeclareCaptionStyle{ruled}{labelfont=normalfont,labelsep=colon,strut=off} %
\floatstyle{ruled}
\newfloat{listing}{tb}{lst}{}
\floatname{listing}{Listing}
\newcommand{\vx}{\mathbf{x}}    %
\newcommand{\vu}{\mathbf{u}}    %
\newcommand{\vf}{\mathbf{f}}    %
\newcommand{\seqX}{\mathbf{X}}    %
\newcommand{\seqU}{\mathbf{U}}    %
\newcommand{\sX}{\mathcal{X}}   %
\newcommand{\sU}{\mathcal{U}}   %
\newcommand{\sW}{\mathcal{W}}   %
\newcommand{\sB}{\mathcal{B}} %
\newcommand{\sM}{\mathcal{M}}   %
\newcommand{\sT}{\mathcal{T}}   %
\newcommand{\sH}{\mathcal{H}}   %

\DeclareMathOperator{\step}{step}
\newtheorem{example}{Example}
\newtheorem{theorem}{Theorem}
\newtheorem{remark}{Remark}
\newtheorem{proposition}{Proposition}

\algrenewcommand\algorithmicindent{1.0em}

\newcommand{\Input}[1]{\Statex \textbf{input:} #1}
\renewcommand{\Output}[1]{\Statex \textbf{output:} #1}
\newcommand{\Parameters}[1]{\Statex \textbf{params:} #1}
\title{db-LaCAM: Fast and Scalable Multi-Robot Kinodynamic Motion Planning with Discontinuity-Bounded Search and Lightweight MAPF}

\author{Akmaral Moldagalieva$^1$, Keisuke Okumura$^{2,3}$, Amanda Prorok$^2$, Wolfgang Hönig$^{1,4}$}
\affiliations{
$^1$Technical University of Berlin, Germany
\\
$^2$University of Cambridge, UK
\\
$^3$National Institute of Advanced Industrial Science and Technology (AIST), Japan
\\
$^4$Robotics Institute Germany, Germany
\\
moldagalieva@tu-berlin.de, \{ko393,asp45\}@cst.cam.ac.uk, hoenig@tu-berlin.de
}

\begin{document}

\maketitle

\begin{abstract}
State-of-the-art multi-robot kinodynamic motion planners struggle to handle more than a few robots due to high computational burden, which limits their scalability and results in slow planning time.
  In this work, we combine the scalability and speed of modern multi-agent path finding (MAPF) algorithms with the dynamic-awareness of kinodynamic planners to address these limitations.
  To this end, we propose discontinuity-Bounded LaCAM (db-LaCAM), a planner that utilizes a precomputed set of motion primitives that respect robot dynamics to generate horizon-length motion sequences, while allowing a user-defined discontinuity between successive motions.
  The planner db-LaCAM is resolution-complete with respect to motion primitives and supports arbitrary robot dynamics.
  Extensive experiments demonstrate that db-LaCAM scales efficiently to scenarios with up to $50$ robots, achieving up to ten times faster runtime compared to state-of-the-art planners, while maintaining comparable solution quality.
  The approach is validated in both 2D and 3D environments with dynamics such as the unicycle and 3D double integrator.
  We demonstrate the safe execution of trajectories planned with db-LaCAM in two distinct physical experiments involving teams of flying robots and car-with-trailer robots.
  
\end{abstract}

\section{Introduction}
Kinodynamic motion planning addresses the problem of finding collision-free trajectories that are dynamically feasible between start and goal states.
This formulation makes it more challenging than geometric planning since it requires accounting for robot dynamics and actuation limits.
For instance, when car dynamics are neglected, the resulting state sequences may be physically infeasible. 
A car cannot rotate in place and must translate while steering to change its heading, making arbitrary state transitions impossible.

In multi-robot settings, the planner must not only enforce each robot's dynamics, but also ensure collision avoidance among robots.
To date, various approaches have been proposed to incorporate robot dynamics into multi-robot planning~\cite{kcbs, cbs-mpc, db-cbs}.
Although these approaches can handle robots with complex dynamics, they remain computationally expensive and do not scale well to large robot teams.

In parallel, multi-robot motion planning, in its simplified form, can be formulated as a Multi-Agent Path Finding (MAPF) problem.
In the MAPF domain, the world is represented as a graph, where robots move between vertices in one step.
Over the years, remarkable progress has been achieved in developing efficient and scalable solutions to MAPF problems~\cite{cbs, ecbs, pibt, okumura2023lacam}.
While these methods can handle thousands of robots~\cite{okumura2023lacam, lacam-star}, they neglect robot dynamics, modeling robots as 2D points that move in a discrete grid world.
As a result, planned trajectories are not executable on real robot platforms.

\begin{figure*}[!t]
    \centering
    \includegraphics[width=\textwidth]{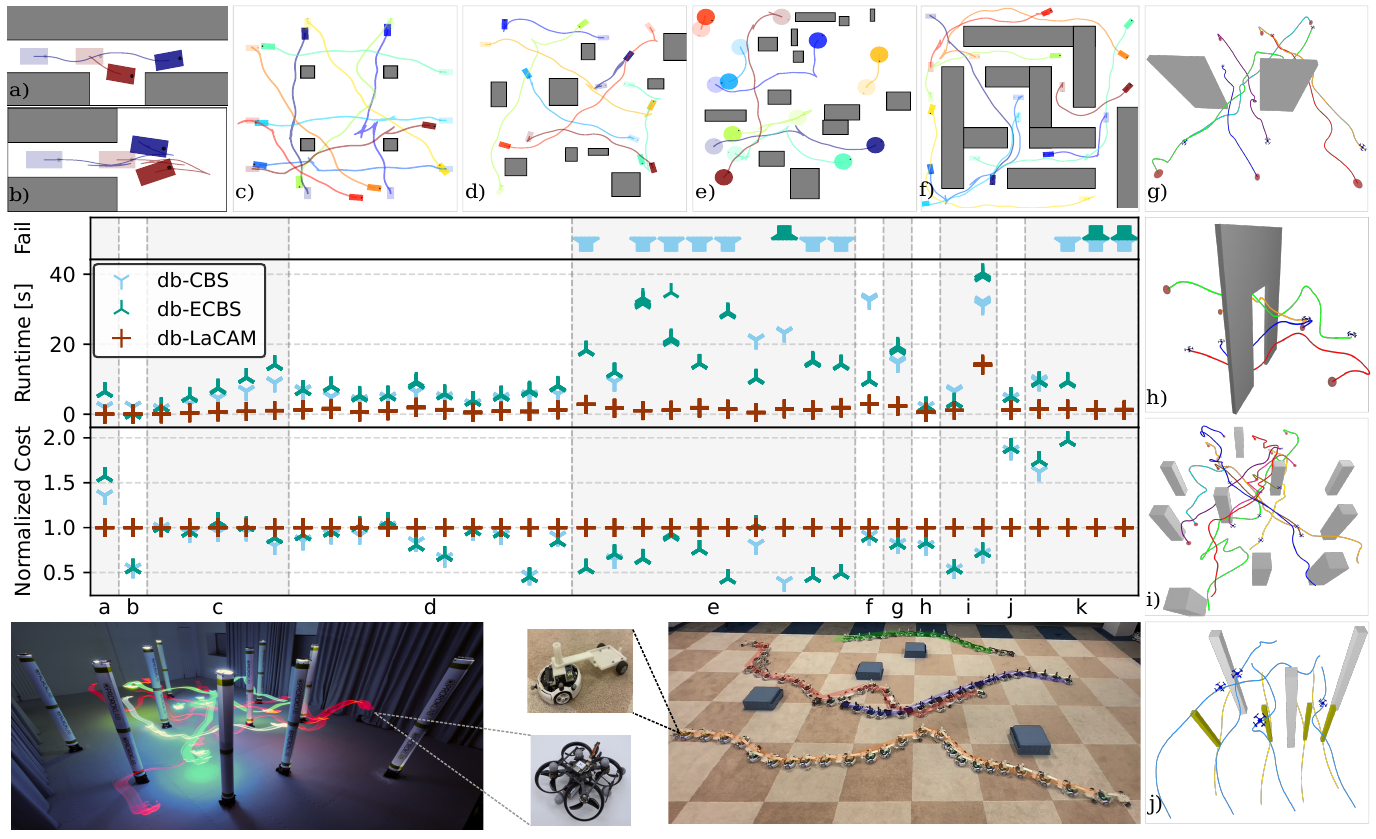}
    \caption{Performance and demonstration of db-LaCAM. Example problem setups (top and third rows) and corresponding quantitative comparisons (second row) over 35 instances grouped into ten representative environments (alternating shaded regions) over 10 trials. Each point represents a single trial outcome, while panels show failure rate, runtime, and normalized cost, respectively. Labels denote the evaluated environments: (a) \emph{alcove}, (b) \emph{at goal}, (c) \emph{circle-n10}, (d) \emph{random-n8}, (e) \emph{random-n8-u$_s$}, (f) \emph{maze-n10}, (g) \emph{passage-n6}, (h) \emph{door-n4}, (i) \emph{forest-n10}, (j) \emph{swap-hetero-n8}, (k) \emph{random-hetero-n8}. Example \emph{random-hetero-n8} denotes a \emph{random} problem with a heterogeneous team of eight robots. Instances (g–i) correspond to 3D environments. The bottom row illustrates real-world demonstrations with a team of ten flying robots and four car-like robots with trailers.}
    \label{fig:main}
\end{figure*}

This work integrates the scalability and efficiency of MAPF algorithms with the dynamic-awareness of kinodynamic planners.
At its core lies the discontinuity-bounded principle~\cite{hoenig2022dbAstar}, which relaxes strict continuity between consecutive motion primitives. 
Allowing bounded discontinuity when selecting successor motions makes the search tractable by enabling primitive reuse and maintaining a finite set of states.
By combining this idea with lightweight MAPF coordination, we propose discontinuity-Bounded LaCAM (db-LaCAM), a dynamics-agnostic multi-robot motion planner capable of generating dynamically feasible and collision-free trajectories efficiently.

The proposed planner db-LaCAM is a search-based planner that extends LaCAM~\cite{okumura2023lacam} to the continuous domain.
As LaCAM is built upon PIBT \cite{pibt}, db-LaCAM incorporates its dynamic counterpart, db-PIBT, to produce feasible motion sequences over a fixed-length horizon.
This integration enables db-LaCAM to leverage db-PIBT for efficient local coordination while performing long-horizon search to avoid livelocks and improve scalability.
The heuristic guidance within db-LaCAM relies on a hierarchical Expansive Space Tree (EST)~\cite{est} that approximates the cost-to-go in continuous, dynamically-aware settings.

Theoretically, db-LaCAM is resolution-complete with respect to motion primitives. 
Empirically, we show that db-LaCAM can solve challenging multi-robot motion planning problems efficiently, see~\cref{fig:main}. 
For instance, db-LaCAM solves a $50$ unicycle robots-instance in $20~\si{s}$, whereas state-of-the-art planners db-CBS~\cite{db-cbs}, db-ECBS~\cite{moldagalieva2025dbecbs} fail.

\section{Related Work}
\label{sec:related_work}
This section reviews related work in multi-agent path finding (MAPF) and multi-robot kinodynamic motion planning. 
\paragraph{Multi-Agent Path Finding}
MAPF assumes a discrete state space represented as a graph. 
A robot can move from one vertex along an edge to an adjacent neighbor in one step; robots cannot occupy the same vertex or traverse the same
edge at the same time step.
Solving MAPF optimally is NP-hard~\cite{yu_2013}; therefore, optimal solvers struggle with scalability~\cite{cbs}.
Several alternative methods have been suggested to address this scalability issue, including suboptimal alternative methods~\cite{ecbs, pibt, okumura2023lacam}.
Existing MAPF solvers are highly scalable and provide solutions efficiently, yet they ignore robot dynamics.
As a result, their solutions can be infeasible to deploy on real robot platforms.

\paragraph{Multi-Robot Kinodynamic Motion Planning}
For multi-robot kinodynamic motion planning, single-robot planners applied to the joint space can be used~\cite{sst}, but do not scale well beyond a few robots.
Better scalability can be achieved by adapting MAPF optimal solvers~\cite{cbs} to kinodynamic motion planning by combining them with Model Predictive Control~\cite{cbs-mpc}, integrating them with sampling-based planners~\cite{kcbs}, or incorporating motion primitives~\cite{db-cbs, moldagalieva2025dbecbs, lf, pibt_ackermann}. 
In the latter case, hand-designed primitives may fail to generalize to complex robot dynamics and limit feasible motions, which can reduce planning efficiency~\cite{pibt_ackermann}.
Additionally, these methods are computationally expensive, leading to slow runtimes.
An efficient planner can be obtained by combining  Mixed-Integer Linear Programs with control theory~\cite{s2m2} or with Bezier curve-based methods~\cite{bezier-curve-planning}.
The former requires safe regions around trajectories for tracking, which can be overly conservative and lead to incompleteness. 
The latter is limited to differentially flat systems.
Overall, existing multi-robot kinodynamic planners remain slow and scale poorly as the number of robots increases.
\section{Preliminaries}
\label{sec:preliminaries}
In this section, we first formally define the problem and its objective considered in this work (\cref{subsec:problem_definition}).
We then briefly review MAPF algorithms that form the basis of our approach (\cref{subsec:background}) and define the notion of a motion primitive, a key component of our planning framework (\cref{subsec:motion_primitive}).
\subsection{Problem Definition} 
\label{subsec:problem_definition}
We consider a team of $N$ heterogeneous robots. 
The state of the $i^{\text{th}}$ robot is given as $\vx^{(i)} \in \sX^{(i)} \subset \mathbb R^{d_{x^{(i)}}}$, which is actuated by controlling actions $\vu^{(i)} \in \sU^{(i)} \subset \mathbb R^{d_{u^{(i)}}}$. 
The workspace the robots operate in is $\sW \subseteq \mathbb R^{d_w}$ ($d_w\in\{2,3\}$), the collision-free space is $\sW_{\mathrm{free}} \subseteq \sW$.

We assume that each robot $i \in \{1,\ldots,N\}$ has dynamics $\dot \vx^{(i)} = \vf^{(i)}(\vx^{(i)}, \vu^{(i)})$.
With zero-order hold discretization, motion can be framed as
\begin{equation}
    \label{eq:dynamics_discrete}
    \vx_{k+1}^{(i)} \approx \step(\vx_k^{(i)}, \vu_k^{(i)}) \equiv \vx_k^{(i)} + \vf^{(i)}(\vx^{(i)}_k, \vu_k^{(i)})\Delta t,
\end{equation}
 where $\Delta t$ is sufficiently small to ensure that the Euler integration holds.

 We denote by $\seqX^{(i)} = \langle \vx_0^{(i)}, \vx_1^{(i)}, \ldots, \vx_{K^{(i)}}^{(i)} \rangle$ the sequence of states of the $i^{\text{th}}$ robot sampled at times $0, \Delta t, \dots, K^{(i)} \Delta t$ and by $\seqU^{(i)} = \langle \vu_0^{(i)}, \vu_1^{(i)}, \ldots, \vu_{K^{(i)}-1}^{(i)} \rangle$ the sequence of actions applied to the $i^{\text{th}}$ robot for times $[0,\Delta t), [\Delta t, 2\Delta t), \ldots, [(K^{(i)}-1)\Delta t, K^{(i)}\Delta t)$.
 
 Our goal is to move a team of $N$ robots from their start states $\vx_s^{(i)} \in \sX^{(i)}$ to their goal states  $\vx_g^{(i)} \in \sX^{(i)}$ as fast as possible while avoiding collisions and respecting robot dynamics.
This problem can be formulated as:
 \begin{align}
    &\min_{\langle\seqX^{(i)}\},\{\seqU^{(i)}\},\{K^{(i)}\rangle} \sum_{i=1}^{N} K^{(i)} \label{eq:opt_mrs}
    \\
    &\text{\noindent s.t.}\begin{cases}
    \vx_{k+1}^{(i)} = \step(\vx_k^{(i)}, \vu_k^{(i)}) & \forall i\; \forall k, \\
    \vu^{(i)}_k \in \sU^{(i)}, \,\,\,\, \vx^{(i)}_k \in \sX^{(i)} & \forall i\; \forall k, \\
    \sB^{(i)}(\vx_k^{(i)}) \in \sW_{\mathrm{free}}  & \forall i \; \forall k,  \\
    \sB^{(i)}(\vx_k^{(i)}) \cap  \sB^{(j)}(\vx_k^{(j)}) = \emptyset  & \forall i \neq j\; \forall k, \\
    \vx^{(i)}_0 = \vx_s^{(i)}, \,\,\,\, \vx^{(i)}_{K^{(i)}} = \vx_g^{(i)} & \forall i, \\
    \end{cases} \nonumber
\end{align}
where $\sB^{(i)}: \sX^{(i)} \to 2^\sW$ is a function that maps the state of the $i^{\text{th}}$ robot to a collision shape. 
The objective is to minimize the sum of the arrival times of all robots.

\begin{example}
	\label{ex:unicycle}
	Consider a car with trailer robot with state $\vx = [x,\; y,\; \theta_1,\; \theta_2]^\top \in \mathbb{R}^2\times (S^1)^2$,
where \((x,y)\) is the car position, \(\theta_1\) is the car orientation and \(\theta_2\) is the trailer orientation.
The control is \(\vu=[v,\;\phi]^\top\in\mathcal{U}\subset\mathbb{R}^2\), with linear velocity \(v\) and steering angle \(\phi\).
Let \(L\) be the car wheelbase and \(L_h\) the hitch length.
	The dynamics are \(\dot{\vx} = [v \cos(\theta_1), v \sin(\theta_1), \frac{v}{L} \tan\phi, \frac{v}{L_h} \sin(\theta_1-\theta_2)]^\top\).
\end{example}

\subsection{PIBT and LaCAM}
\label{subsec:background}
PIBT \emph{(priority inheritance with backtracking)}~\cite{pibt} is a scalable and suboptimal MAPF algorithm. 
The method performs one-timestep planning among agents following the priority of the agents. 
In each timestep, agents update priorities, and the planner assigns next positions sequentially to avoid conflicts with high-prioritized agents.
When there is no location left for a lower-prioritized agent, then priority inheritance takes place, enforcing the higher-prioritized robot to move out of the way. 
PIBT is a greedy algorithm, which is guided by cost-to-go heuristics.
Its greedy nature can cause deadlocks, making the planner incomplete.

The complete planner LaCAM \emph{(lazy constraints addition search)}~\cite{okumura2023lacam} addresses these limitations of PIBT.
LaCAM is a search-based MAPF solver that works in two levels. 
At the high level, LaCAM searches over configurations of all agents.
At the low level, it generates constraints for each high-level node. 
Constraints are used to specify which locations are occupied by which agent in the next configuration.
The high-level search is guided by PIBT and proceeds lazily: instead of generating all valid successors, it only generates one feasible successor each time the high-level node is invoked.

The scalability and speed of PIBT and LaCAM motivate extending them to kinodynamic motion planning.
Such an extension introduces new challenges: continuous state spaces, constraints from robot dynamics, and harder heuristic computation, as Euclidean distance no longer reflects true cost-to-go.
The proposed planner, db-LaCAM, addresses these challenges via motion primitives, enabling efficient, constraint-respecting planning in the continuous domain.

\subsection{Motion Primitive}
\label{subsec:motion_primitive}
A motion primitive is a sequence of states and controls that fulfill the dynamics of the system given in~\cref{eq:dynamics_discrete}. Formally,
\begin{definition}
    \label{definition:motion_primitive}
    A motion primitive is a tuple $ \langle \seqX, \seqU, K \rangle$, consisting of state sequences  $\seqX = \langle \vx_0,..., \vx_K \rangle$ and control sequences $\seqU = \langle \vu_0, ..., \vu_{K-1} \rangle$ which obey the dynamics $\vx_{k+1} = \step(\vx_k, \vu_k)$. 
\end{definition}
Motion primitives can be generated by discretizing the state space of the robot~\cite{lf, primitive-swarm}, or by solving a two-point boundary value problem with nonlinear optimization~\cite{hoenig2022dbAstar}; this work employs the latter.

\section{Approach}
\label{sec:approach}
We begin by describing db-PIBT (\cref{subsec:dbpibt}), the core component of our planner db-LaCAM.
We then present db-LaCAM (\cref{subsec:dblacam}), followed by the techniques developed for heuristic estimation (\cref{subsec:heuristics}), motion clustering (\cref{subsec:motion_clustering}), livelock detection and resolution (\cref{subsec:livelock}), and finally discuss the properties of db-LaCAM (\cref{subsec:properties}).
\subsection{db-PIBT} 
\label{subsec:dbpibt}
Kinodynamic motion planner db-PIBT extends PIBT~\cite{pibt} to the continuous domain. 
db-PIBT (\cref{alg:dbpibt}) searches over motion primitives to plan a fixed-horizon state sequence and supports arbitrary robot dynamics with pre-computed primitives; major changes are in blue.

At a high level, db-PIBT incrementally assigns dynamically feasible motion segments to robots in descending priority order (Line~\ref{pibtline:call-func}).
The priority order of robots is based on the distance to the goal, with the farthest receiving the highest priority. 
Here, we rely on a user-defined \emph{metric} $d: \sX \times \sX \to \mathbb R$ to measure the distance between two states.

The planner takes as input $N$ robots and a list of processed motion sets $\boldsymbol{\hat{M}} = (\hat{\sM^1},\dots,\hat{\sM^N})$.
Within the high-level loop, a recursive procedure is called for each robot.
The procedure \texttt{db-PIBT} explores the set of valid motion primitives $\hat{\sM^i}$ for the given robot $i$ (Line~\ref{pibtline:for_loop}) to select feasible successors.
Each candidate motion $m$ is checked for potential conflicts:
\emph{(i)} collisions with high-priority robot motions (Line~\ref{pibtline:collision_planned}); \emph{(ii)} collisions with lower-priority robots that are not yet planned
(Line~\ref{pibtline:collision_unplanned}).
For the second case, we assume that each lower-priority robot $j$ has a set of valid motion primitives $\hat{\sM^j}$ (Line~\ref{pibtline:potential_motions}). 
When the candidate motion $m$ conflicts with any motion in $\hat{\sM^j}$, then db-PIBT recursively invokes itself (Line~\ref{pibtline:call_dbpit}) to tentatively assign a valid motion to robot $j$.
This recursive call checks whether a consistent set of motions can be found for both robots.
If no valid combination exists, the candidate motion $m$ is discarded, and the algorithm proceeds to test the next motion in $\hat{\sM^i}$.
Once a valid motion primitive is found, db-PIBT updates the reserved motions vector $\sT$ with the found motion for robot $i$ (Line~\ref{pibtline:reserve_motion}) and returns \texttt{VALID} (Line~\ref{pibtline:return_valid}).

While db-PIBT is effective, its greedy nature can lead to livelocks in tightly constrained scenarios, highlighting the need for long-term planning, which db-LaCAM provides.

\renewcommand{\hl}[1]{\textcolor{blue}{#1}} 
\begin{algorithm}[ht]
\caption{db-PIBT}
\label{alg:dbpibt}
\begin{algorithmic}[1]
\Input{robots $N$, list of processed motion sets \hl{$\boldsymbol{\hat{M}}$}}
\Output{motions \hl{$\sT$} (each element is initialized with $\bot$)}
\medskip
\State \textbf{for}~$i \in N$~\textbf{do};~
             \textbf{if}~{\hl{$\sT[i]$} $ = \bot$}~\textbf{then}~{\hl{db-PIBT$(i, \boldsymbol{\hat{M}}, \sT)$}}
      \label{pibtline:call-func}
      \State \Return $\sT$
\Statex
\Procedure{\texttt{db-PIBT}}{$i,\boldsymbol{\hat{M}},\sT$}      
\For{\hl{$m \in \hat{\sM^i}$}} \Comment{loop over sorted motions} \label{pibtline:for_loop}
 \If{\hl{$\exists\, m_k \in \sT$ \textbf{s.t.}\ \texttt{Collide}$(m, m_k)$}} \label{pibtline:collision_planned}
 \State \textbf{continue} \Comment{collision with planned robot}  \EndIf 
 \State \hl{$\sT[i] \gets m$} \Comment{reserve the motion} \label{pibtline:reserve_motion}
 \For{\hl{$\hat{\sM^j} \in \boldsymbol{\hat{M}}$}}\label{pibtline:potential_motions}
 \If{\hl{$\exists\, m_j \in \hat{\sM^j}$ \textbf{s.t.}\ \texttt{Collide}$(m, m_j)$}} \label{pibtline:collision_unplanned}
\If{\hl{db-PIBT($j,\boldsymbol{\hat{M}},\sT$)} = \text{INVALID}} \label{pibtline:call_dbpit}
    \State \textbf{continue} \Comment{priority-inheriting robot failed}   \EndIf
\EndIf
\State \Return VALID \label{pibtline:return_valid}
\EndFor
\EndFor
\State \Return INVALID
\EndProcedure
\end{algorithmic}
\end{algorithm}

\subsection{db-LaCAM}
\label{subsec:dblacam}
The planner db-LaCAM is a search-based kinodynamic planner that builds on LaCAM~\cite{okumura2023lacam} and db-PIBT to account for robot dynamics.
The high-level search of db-LaCAM is given in~\cref{alg:dblacam}.
The algorithm can work with arbitrary robot dynamics.
Its major changes compared to LaCAM are marked in blue.

The planner db-LaCAM starts the search by creating two sets: $Open$ that stores nodes to be expanded, and $Explored$ to keep track of already expanded nodes (Line~\ref{line:init}).
The initial node, given as $Q_{init}$, is initialized with a set of starting states for all robots, an empty set of constraints, and motion primitives (Line~\ref{line:init_node}).
At each iteration, the top node from $Open$ is removed and expanded.
If the state of this node is within the user-defined $\delta_g$ distance to goal states, the solution is recovered by backtracking each node's stored motion and returned (Line~\ref{line:reach_goal}).
Otherwise, the search proceeds by querying a set of valid motion primitives via \texttt{Process\_Motions}.
Motions generated with \texttt{Process\_Motions} update the constraint tree of the current node $Q$ with \texttt{Set\_Constraint\_Tree} (Line~\ref{line:update_constraints}).
Each constraint specifies the motion of each robot over the subsequent planning horizon (details are in the appendix).
The \texttt{db-PIBT} then assigns priorities to robots and sequentially plans for each robot
using the updated constraint tree to guide motion selection and maintain inter-robot consistency.
The output of \texttt{db-PIBT} is a next horizon, collision-free sequence of states for each robot (Line~\ref{line:motion_generator}).
If all robots move successfully, then this configuration creates a new high-level node $Q'$ (Line~\ref{line:q_new}) and adds it to the $Open$ set (Line~\ref{line:openset_add}).
After each successful high-level search iteration, the robots are optionally checked for livelock (Line~\ref{line:livelock}).

\begin{figure}[t]
  \hfill 
  \includegraphics[width=1.0\linewidth]{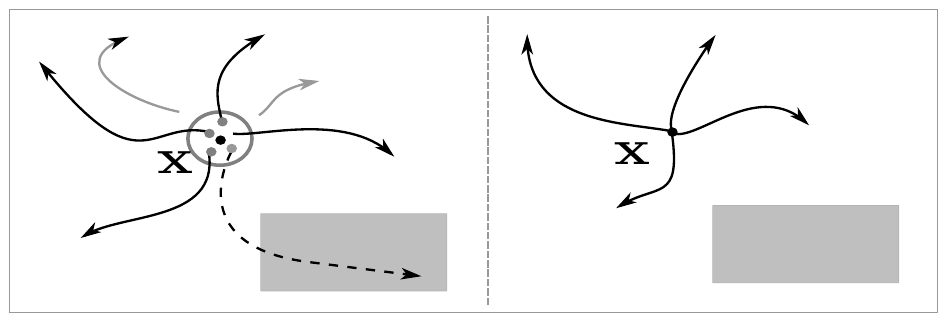}
  \caption{Visual representation of motion primitive samples. Given state $\vx$, applicable motion primitives (black edges) start within a discontinuity lower than $\alpha \delta$ (gray circumference). Action sequences of applicable motion primitives forward-propagate the state $\vx$; motions are discarded (dashed edges) if any state is in collision.}
   \label{fig:motions}
\end{figure}

The \texttt{Process\_Motions} procedure is a key component of the algorithm and also its most computationally expensive step.
It involves several subroutines for generating, validating, and ranking motion candidates as shown in~\cref{alg:dblacam}.
First, it finds applicable motions.
A motion $m$ is considered applicable at state $\vx$ if its start state $m.\vx_s$ is within $\alpha \delta$ away from it, where $\alpha$ is a user-defined parameter and $\delta$ is the discontinuity bound (Line~\ref{line:choose_applicable_motions}).
For efficient search, we adopt a $k$-d tree, $\sT_m$, to index the start states of all provided motion primitives (Line~\ref{line:init_tm}).
Once applicable motions are selected, their action sequences are applied to the current states to perform forward propagation (Line~\ref{line:rollout_motions}). 
An example of resulting motions is illustrated in ~\cref{fig:motions}. 
Second, it computes the cost-to-go $h(m.\vx_f)$ for each rolled-out motion $m$ (Line~\ref{line:assign_heuristics}); details are in~\cref{subsec:heuristics}.
Finally, it performs a motion clustering (Line~\ref{line:cluster_motions}), as described in~\cref{subsec:motion_clustering}. 

\renewcommand{\hl}[1]{\textcolor{blue}{#1}} %
\begin{algorithm}[t]
\caption{db-LaCAM}
\label{alg:dblacam}
\begin{algorithmic}[1]
\Input{robots $N$, start states $\langle\vx_s^{(i)}\rangle$, goal states $\langle\vx_g^{(i)}\rangle$, list of motion primitive sets $\boldsymbol{M} = (\sM^1,\dots,\sM^N)$}
\Parameters{goal threshold $\delta_g$}
\medskip
\State{$\sT_m \gets \texttt{Nearest\_Neighbor\_Init}(\boldsymbol{M})$} \Comment{use start states of motion primitives} \label{line:init_tm}
\State \texttt{Initialize} $Open$, $Explored$ \label{line:init}
\State{$C_{\text{init}} \gets \langle~parent:\bot, who:\bot, where:\bot, \hl{motion : \{ \}}~\rangle$} \Comment{no constraint}
\State $Q_{\text{init}} \gets \langle state :\vx_s,\ tree :C_{\text{init}}, \ \hl{motions : \{ \}} \rangle$ \label{line:init_node}
\State $Open.push(Q_{\text{init}})$; $Explored[\vx_s] \gets Q_{\text{init}}$
\While{$Open \neq \emptyset$}
\label{algo:star:while-start}
  \State $Q \gets Open.top()$ \Comment{high-level node} \label{line:pop_node}
  \State $\vx \gets Q.state$ \Comment{starting state for the horizon}
  \If{\texttt{distance}$(\vx, \vx_g) \leq \delta_g$}
    \State \Return backtrack($Q$) \EndIf \label{line:reach_goal}
    \If{$Q.tree = \emptyset$} $Open.pop()$; \textbf{continue} \EndIf
  \State $C \gets Q.tree.pop()$ \Comment{constraint tree}
  \State \hl{$\boldsymbol{\hat{M}} \gets$ \texttt{Process\_Motions}($\vx, \vx_g, \sT_m$)} \label{line:get_motions}
  \State  \hl{\texttt{Set\_Constraint\_Tree}($\boldsymbol{\hat{M}}, Q, C$)} \Comment{\cref{alg:update_constraint_tree}}\label{line:update_constraints}
  \State \hl{$\sT' \gets \texttt{db-PIBT}(N,\boldsymbol{\hat{M}})$} \label{line:motion_generator} \Comment{\cref{alg:dbpibt}}
  \State \hl {$\vx' \gets \sT'.\vx.back()$} \Comment{final state for the horizon}
  \If{$\vx' = \bot$} \textbf{continue} \EndIf
  \If{$Explored[\vx'] \neq \bot$} \textbf{continue} \EndIf
  \State $Q' \gets \langle state: \vx',\ tree: C_{init}, \ \hl{motions: \sT'} \rangle$  \label{line:q_new}
  \State $Open.push(Q')$; $Explored[\vx'] \gets Q'$ \label{line:openset_add}
  \State \hl{\texttt{Livelock\_Detection(Q)}} \label{line:livelock} \Comment{optional step}
\EndWhile
\State \Return NO\_SOLUTION
\Statex
\Procedure{process\_motions}{$\vx, \vx_g,\sT_m$}
  \For{$i \in N$}
    \State{$\sM^i_a \gets$ \texttt{Nearest\_Neighbor\_Query($\sT_m^i, \vx^i$)}}
    \Comment{applicable motions with discontinuity up to $\alpha \delta$}
    \label{line:choose_applicable_motions} 
   \State{$\sM_r^i \gets$ \texttt{Rollout\_Motions($\vx^i, \sM^i_a$)}}\label{line:rollout_motions} 
    \State{\texttt{HEST($\vx_g^i, \sM_r^i$)}} \label{line:assign_heuristics}
    \Comment{compute heuristics,~\cref{subsec:heuristics}}
   \State{$\hat{\sM}^i \gets $ \texttt{Cluster\_Motions($\sM_r^i$)}}\label{line:cluster_motions} \Comment{~\cref{subsec:motion_clustering}}
   \EndFor
   \Return $\hat{\sM}$ \Comment{clustered motions for each robot}
\EndProcedure
\end{algorithmic}
\end{algorithm}

\subsection{Heuristics Estimation} 
\label{subsec:heuristics}
The core component of db-LaCAM, db-PIBT, requires an accurate estimation of the cost-to-go $h$ for each motion to guide the search towards the goal.
In the discrete domain, PIBT uses the shortest path length to compute the $h$ for each grid cell. 
However, in a continuous domain where robots exhibit different dynamics, this metric can no longer provide a reliable $h$.
The $h$ can be pre-computed using a single robot planner in a reverse manner from the goal state towards the start state.
This method has two limitations. 
First, it requires inverse dynamics, which are often intractable to compute since the system is forward-propagated via actions, and recovering actions backward from the final state would need inverting the dynamics model.
Second, it is computationally expensive -- especially in large environments, as it explores large, often irrelevant regions of the environment.

\paragraph{Hierarchical EST}
We propose the Hierarchical Expansive Spaces Trees (\emph{HEST}) for efficient estimation of $h$.
\emph{HEST} adapts Guided EST~\cite{est}, a single-robot sampling-based planner that expands the most promising state 
until reaching the goal.

\emph{HEST} implements a nearest-neighbor table of explored states with their cost-to-go values for efficient $h$-value retrieval during expansions.
It operates on two hierarchical levels.
At the high level, it runs Guided EST in reverse from the goal to the start to obtain a coarse estimate of the heuristic $h$. 
At the low level, it conducts a forward search with Guided EST from a given state towards the goal, using the high-level estimate to prioritize expansions. 

\cref{alg:hierarchical_est} computes motion-wise heuristic estimates for robot $i$ using precomputed nearest-neighbor tables.
For each motion in $\sM^i_r$ (Line~\ref{estline:for_loop}), the algorithm queries the reverse search table $H_r^i$ to find the closest stored state (Line~\ref{estline:nearest}).
If no nearby state exists within threshold $\Delta$, it performs a forward estimation using $\texttt{EST}$ (Line~\ref{estline:est_forward}) and updates the forward table $H^i$; otherwise, it reuses the nearest neighbor’s heuristic value (Line~\ref{estline:assign_h}).
This structure enables \emph{HEST} to focus computation where it matters most--refining heuristics locally while reusing information from high-level exploration.
An ablation study in~\cref{sec:ablation_study} examines heuristic methods.

\begin{algorithm}[ht]
\caption{HEST for a single robot $i$}
\label{alg:hierarchical_est}
\begin{algorithmic}[1]
\Input{goal state $\vx_g^i$, rolled-out motion set $\sM^i_r$}
\Output{heuristic values $h$ for each motion in $\sM^i_r$}
\Parameters{distance threshold $\Delta$}
\Statex \textbf{global:} $H_r^i, H^i$ \Comment{Nearest-neighbor table for reverse and forward searches}
\Statex
\For{$m \in \sM^i_r$} \label{estline:for_loop}
  \State{$\vx_n \gets \texttt{Nearest\_Neighbor\_Query}(m.\vx_f, \sH_r^i)$} \label{estline:nearest}
  \If{$\vx_n = \emptyset \ \lor\ \texttt{distance}(m.\vx_f, \vx_n) > \Delta$}
    \State{$h(m.\vx_f) \gets \texttt{EST}(m.\vx_f, \vx_g, \sH^i)$} \Comment{update $H^i$}
    \label{estline:est_forward}
  \Else
    \State{$h(m.\vx_f) \gets h(\vx_n)$} \label{estline:assign_h} \Comment{assign nearest neighbor's $h$}
  \EndIf
\EndFor
\end{algorithmic}
\end{algorithm}

\subsection{Motion Primitives Clustering}
\label{subsec:motion_clustering}
\begin{figure}[t]
  \includegraphics[width=1.0\linewidth]{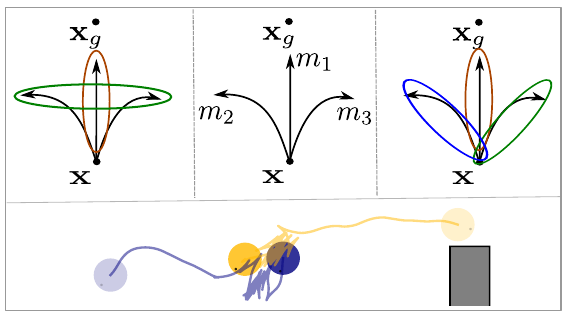}
  \caption{Top row: motion primitive clusters — middle: original motions, left: GOC, right: SC-GOC, with colors for each cluster. Motions toward left and right have similar distance to the goal state, yielding close $h$-values. Bottom row: example livelock between two robots.}
   \label{fig:clustering}
\end{figure}

Let $\sM^i = \langle \, m_1,\; m_2,\; \dots \, m_M\rangle$ be a set of valid motions for the robot $i$. 
As the number of motions in $\sM^i$ grows, the search in db-PIBT becomes less efficient and slows down.
A straightforward approach to reduce the size is to sort motions in ascending order of their heuristic $h$ values and select the top $n$ motions.
However, this can lead to a set of motions that are very similar to each other, which might hinder the planner.
Based on the motion primitives design, we develop two clustering techniques to have diverse motions.
\paragraph{Goal-Oriented Clustering (GOC)} GOC aims to sort motions in an ascending order of h-values $h_1 \le h_2 \le \dots \le h_M$.
For a cluster $\mathcal{K}$ starting at $h_l$, it includes all motions $m\in\sM_i$ that satisfy $| h - h_l | \le \iota$, where $\iota = \rho \cdot (h_\text{max} - h_\text{min})$ and $\rho$ is a hyperparameter.
Once clusters are formed, $n$ elements per cluster are chosen using one of:  \emph{(i)} vanilla selection--all elements; \emph{(ii)} deterministic selection--top $n$ elements; \emph{(iii)} weighted selection--sampling with probability proportional to their values.
This clustering method speeds goal attainment but can cause livelock when robots must pass near their goals.
As in~\cref{fig:clustering} (bottom), both robots may prefer forward motions with lower $h$, blocking each other.

\paragraph{Space-Cover and Goal-Oriented Clustering (SC-GOC)} SC-GOC aims to provide a set of diverse motions for a better space cover.
It first forms multiple intermediate clusters $\mathcal{K}_{int}$ based on spatial proximity.
Specifically, it picks the motion $m_r$ from $\sM^i$ with the minimum $h$-value and sets it as the reference motion.
Then, it computes the distance between the last states of each motion $m\in\sM^i$ to the $m_r$ as $d(m.\vx_f, m_\text{r}.\vx_f)$.
An intermediate cluster $\mathcal{K}_{int}$ is generated by including all motions that have a distance lower than the threshold $\tau$.
The value of $\tau$ is a hyperparameter.
After, within each cluster, trajectories are sorted by $h$-value, and element selection can be performed using the same strategies \emph{(i-iii)} used for GOC.
Selected $n$ motions are then reordered in an inside-out sequence, starting from the middle element and alternating outward toward the ends:
$\langle\, m_{\lceil n/2 \rceil},\; m_{\lceil n/2 \rceil - 1},\; m_{\lceil n/2 \rceil + 1},\; \dots \,\rangle$. 
Generated motion clusters for GOC (left) and SC-GOC (right) are shown in~\cref{fig:clustering}.
Analysis of db-LaCAM’s performance using different clustering methods is presented in~\cref{sec:ablation_study}.

\subsection{Livelock Behavior}
\label{subsec:livelock}
PIBT suffers from deadlock/livelock due to its short-horizon nature.
One example is illustrated in \cref{fig:clustering} (bottom row).
In MAPF settings, the livelock problem can be resolved by introducing special techniques like \textit{swap}~\cite{lacam-star}, 
or by backtracking recent agent-wise state histories~\cite{jain2025lagat}.
Using the same techniques in the continuous domain is not trivial.
For instance, a unicycle robot might return to the same position with a different orientation, which is treated as a new state distinct from previously visited states.

\paragraph{Detection and Recovery}
We resolve livelock as follows: \emph{(i)} identifying robots that have oscillations in their motions; \emph{(ii)} using state-aware motion clustering to provide a better space cover. 
For \emph{(i)}, each time a new motion is generated in~\cref{alg:dblacam}, each robot's heuristic history is backtracked to check for oscillations. 
A robot is considered to be in a livelock if the change $\Delta h$ between consecutive high-level nodes alternates in sign.
Once robots are identified, we break the livelock behavior by providing more diverse motions to the search with SC-GOC.
The intuition is that with a diverse set of motions—including different orientations—robots are more likely to take sideway motions, effectively breaking potential livelocks.
As discussed in~\cref{subsec:motion_clustering}, SC-GOC provides a good cover of the space by clustering motions based on spatial proximity.

A visual representation of the final clustering results using GOC and SC-GOC is given in~\cref{fig:clustering} (top row).
Here, although $m_1$ is the best option to reach the goal, it triggers livelock unless one of the robots decides to move sideways.
However, sideway motions ($m_2, m_3$) have a higher $h$-value, so can never be taken by db-PIBT.
Moreover, with the GOC clustering technique, motions $m_2$ and $m_3$ can be grouped in one cluster due to their similar distances to the goal state $\vx_g$.
In \cref{sec:ablation_study}, we analyze motion clustering methods. 

\subsection{Properties}
\label{subsec:properties}

The planner db-LaCAM is probabilistic resolution-complete (PRC; for any fixed resolution, the probability of finding a solution, if one exists at that resolution, approaches 1). 
\begin{theorem}
    \label{theorem:ao}
    The db-LaCAM motion planner in \cref{alg:dblacam} is probabilistically resolution-complete when elements in each cluster are selected probabilistically based on their weights.
\end{theorem}
\begin{proofsketch}
    Db-LaCAM is probabilistically complete up to the resolution of the implicit graph defined by the chosen motion primitives.
    With an exhaustive search over the finite search space, it eventually finds a solution if one exists.
    Because motion clustering involves stochastic sampling, there exists a non-zero probability that db-LaCAM selects the correct motion.
\end{proofsketch}

We consider the time complexity of db-PIBT in~\cref{alg:dbpibt}.
\begin{proposition}
    \label{remark:pibt_complexity}
    The time complexity of db-PIBT for a single fixed-length horizon planning is $\mathcal{O}(N^2M^2)$, where $N$ is the number of robots, $M$ is the maximum number of motion primitives.
\end{proposition}
Details of Theorem 1 and the proof of Proposition 1 are provided in the appendix.

\section{Experimental Evaluation}
\label{sec:validation}

For benchmarking we consider robot dynamics like unicycle ($1^\text{(st)}$-order), double integrator 3D.

We test db-LaCAM with discontinuity-bounded CBS (db-CBS)~\cite{db-cbs} and its suboptimal variant db-ECBS~\cite{moldagalieva2025dbecbs}.
We analyze the success rate, computational time until the first solution is found, solution cost.
The cost is a time, equal to the sum of control duration over all single-robot paths.

Our planner db-LaCAM is implemented in C++.
For db-CBS and db-ECBS, we use the respective publicly available implementations from the authors.
All planners rely on FCL (Flexible Collision Library)~\cite{FCL} for collision checking.
The experiments are run on a workstation (AMD Ryzen Threadripper PRO 5975WX @ 3.6 GHz, 64 GB RAM, Ubuntu 22.04).
The code and problem instances are publicly available\footnote{\url{https://github.com/IMRCLab/db-lacam}}.

\subsection{Benchmarking}
Motion primitives are pre-computed offline by solving two-point boundary value problems with random start and goal configurations with nonlinear optimization~\cite{ortiz2024}.
Motion clustering uses deterministic selection to choose elements per cluster.
The used hyperparameter values are listed in the appendix.
An instance is not solved successfully if no solution is found after the timelimit.

\paragraph{Environments}
Each environment is modeled as a set of axis-aligned cuboid obstacles, each defined by its
center position and dimensions.
The timelimit for all instances is $60~\si{s}$.
We consider the following scenarios: 
\begin{itemize}
    \item \textbf{Canonical 2D} examples include problems like \emph{alcove, atgoal}, where one of the robots is forced to move to the alcove or aside to let the other robot reach its goal (\cref{fig:main}a--b). 

    \item{\textbf{Circle 2D}} example has from $N=2$ up to $N=10$ unicycle robots operating in $11\times11$ environment, where robots need to swap to reach their goals (\cref{fig:main}c).
    
    \item \textbf{Random 2D} instances feature randomly placed obstacles and randomly assigned start and goal states for $N=8$ unicycle robots in a $10\times10$ environment.
    Of these, 10 instances use box-shaped robots (\cref{fig:main}d), and the other 10 use spherical robots with a radius of $20~\si{cm}$ (\cref{fig:main}e).

   \item \textbf{Maze} instance includes narrow corridors that force frequent close-proximity passes with $N=10$ unicycle robots (\cref{fig:main}f).
   
    \item \textbf{Problems 3D} comprise compact environments with moderate obstacle density, including narrow corridors and cluttered spaces (\cref{fig:main}g--i).
    The environments measure up to approximately $4 \times 6 \times 1.5$, and the number of robots ranges from four to ten.

   \item{\textbf{Heterogeneous Robots}} problems involve teams of ten robots (e.g., unicycle, 3D double integrator) in environments with and without obstacles. 
   Some instances require robot swapping (\cref{fig:main}j), while others feature random start and goal states.
\end{itemize}

\paragraph{Results}

Overall, db-CBS performs poorly in \emph{2D} environments with frequent close-proximity interactions, such as those involving large or spherical robots, where numerous inter-robot conflicts lead to extended runtimes and low success rates (around $30\%$). 
In contrast, db-ECBS handles these scenarios more effectively, achieving success rates up to $90\%$ by leveraging its conflict-guided heuristic to focus the low-level search. 
The planner db-LaCAM consistently outperforms all other planners, solving every instance rapidly (below \SI{3}{\second} even in the most cluttered setups) and maintaining high success across all environment types.

In \emph{3D} environments, db-LaCAM achieves the lowest runtime consistently across all instances.
Planners db-CBS and db-ECBS perform reasonably well in compact environments such as \emph{forest}, \emph{door}, but their efficiency degrades as the environment size and search space grow.
In larger 3D setups like \emph{passage}, both planners often fail to compute solutions within the timelimit due to the high computational cost of the reverse-search heuristic.

With \emph{heterogeneous robots} instances, db-LaCAM consistently solves all problems with the lowest runtime. 
Planners db-CBS and db-ECBS achieve only around $60\%$ success, as the optimization component often fails to converge, though the discrete search produces a solution.

\paragraph{Physical Robots}
We demonstrate the safe execution of db-LaCAM trajectories on real robot platforms.
The real-world experiments are conducted inside a $7 \times 4 \times 2.5\,\si{m^3}$ room equipped with a motion capture system with twelve Optitrack cameras. 
We use Sanity custom drones for flying robots~\cite{woo2025sanity}, and Pololu 3pi+ 2040 differential-drive robots for ground robots.
Pictures of the deployed platforms are illustrated in~\cref{fig:main} (bottom row).
In the first scenario, we test the \emph{forest} example with ten flying robots modeled as 3D double integrator dynamics. 
In the second scenario, four ground robots modeled as cars with trailers are tasked to swap their positions.

\subsection{Scalability Test}
\begin{figure}[h]
    \centering
    \includegraphics[width=1.0\linewidth]{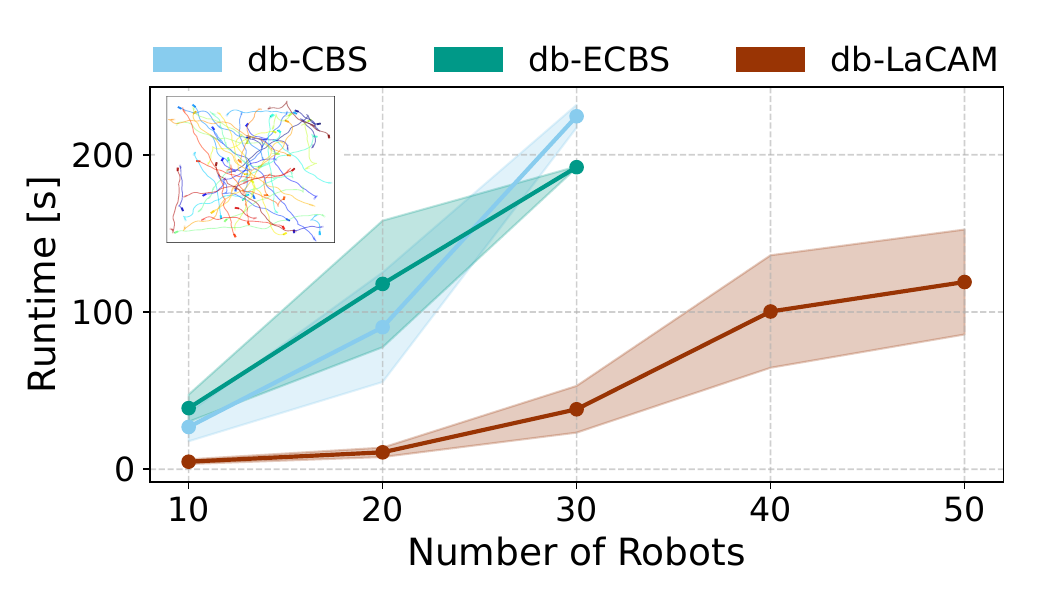}
    \caption{Runtime required to find a feasible solution for different numbers of robots over five trials.}
    \label{fig:scalability}
\end{figure}

We consider teams of $N=10,20,30,40,50$ unicycle robots operating in $20\times20$ size environment (\cref{fig:scalability}).
Start, goal states for each robot are generated randomly in each instance.
We set the timelimit to $5~\si{min}$, since baselines like db-CBS, db-ECBS require a longer time to find a solution for cases with $N > 10$ robots.
    
\paragraph{Results} 
Scalability test results with increasing number of robots are shown in~\cref{fig:scalability}.
The planner db-LaCAM consistently solves all instances with low runtime across different team sizes. 
For large teams ($N\geq40$), both db-CBS and db-ECBS frequently fail, highlighting db-LaCAM’s robustness in dense and challenging environments.

\subsection{Ablation Study}
\label{sec:ablation_study}

\begin{figure}[h]
    \centering
    \includegraphics[width=1.0\linewidth]{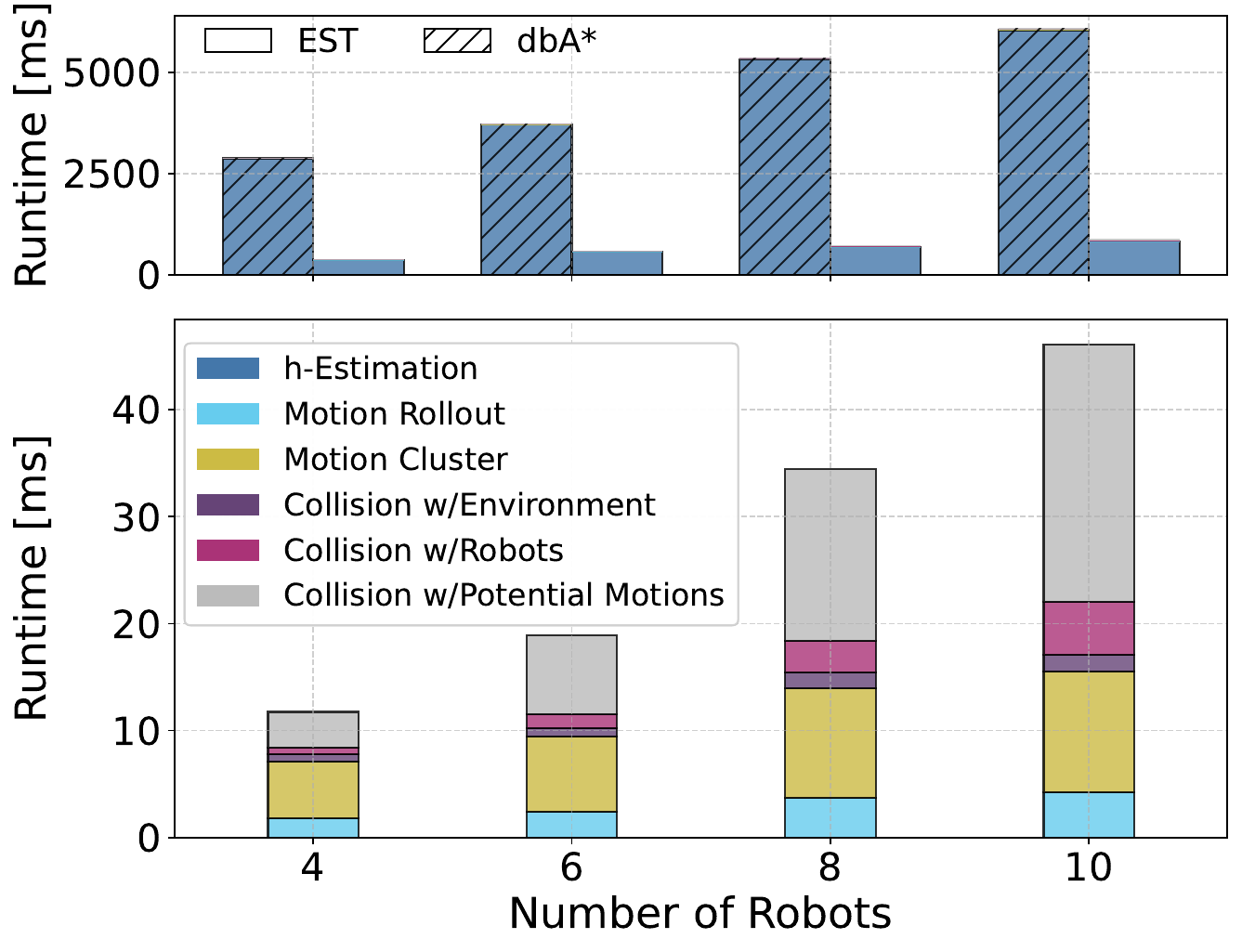}
    \caption{Computation time analysis for db-LaCAM. Upper row: two different methods for the heuristics $h$ computation. Bottom row: analysis of time spent on some key components of db-LaCAM. Experiments are conducted using \emph{circle 2D} example with varying numbers of robots as labeled.}
    \label{fig:ablation_heuristic_estimation}
\end{figure}

\paragraph{Computation time analysis}
We evaluate how much computational time is spent on some key components of db-LaCAM and discuss how it varies with the increasing number of robots.
Time statistics for key components of db-LaCAM while solving the \emph{circle 2D} example are given in~\cref{fig:ablation_heuristic_estimation}.
The total time is mainly occupied by the heuristics $h$ estimation, which will be discussed in detail in the following paragraph.
The second most expensive operation is the collision checking against the potential motions of low-priority neighboring robots that have no plans yet.
Motion clustering is another time-consuming component, though it is relatively insensitive to the number of robots.
Finally, collision with high-priority robots and motion rollout operations exhibit similar runtime across all instances. 

\paragraph{Analysis of Heuristic Look-up Table Estimation}
We discuss two methods for heuristic value $h$ computation from~\cref{subsec:heuristics}: \emph{(i)} a single-robot kinodynamic planner db-A* (Discontinuity-Bounded A*)~\cite{hoenig2022dbAstar} run in a reverse manner from $\vx_g$ to  $\vx_s$; \emph{(ii)} \emph{HEST}.
We analyze how the choice of the heuristic estimation affects the runtime of the planner.
The summary of results is given in~\cref{fig:ablation_heuristic_estimation}.
\emph{HEST}-based heuristic estimation method runs significantly faster compared to db-A*-based reverse search. 
This is due to the db-A*-based method exploring unnecessary parts of the environment as shown in~\cref{fig:ablation_node_expansion} in the appendix. 
Whereas, \emph{HEST}-based method explores only the relevant part of the environment. 
This happens because \emph{HEST} explores nodes on demand, when the lookup table does not contain a nearby state and fails to return a $h$-value.

\begin{figure}[h]
    \centering
    \includegraphics[width=1.0\linewidth]{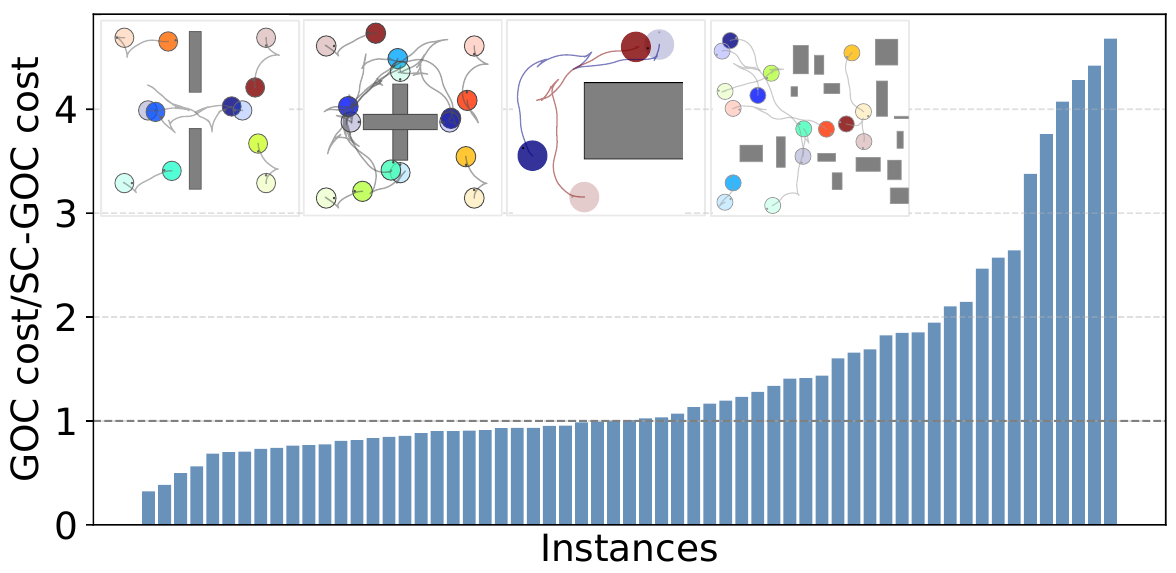}
    \caption{Analysis of two different methods for motion clustering: GOC and SC-GOC. Evaluation is performed with a \emph{random} problem instance with eight unicycle robots of spherical shape with radius $20~\si{cm}$ over $70$ instances with varying settings.}
    \label{fig:ablation_clustering}
\end{figure}

\paragraph{Analysis of Motion Primitive Clustering}
We evaluate two methods for clustering applicable motion primitives discussed in~\cref{subsec:motion_clustering}: \emph{(i)} motions are grouped based on their heuristic value $h$, thus goal reaching is faster (GOC); \emph{(ii)} motions are grouped based on relative distance and heuristic value $h$ enabling a better cover of the space (SC-GOC). 
Both methods employ weighted selection to choose elements from each cluster.
We compare the impact of the two methods on the solution cost of db-LaCAM. 
Results are summarized in~\cref{fig:ablation_clustering}.

Instances are designed to have close-proximity interaction between robots to produce livelock cases.
In these cases, db-LaCAM with GOC performs poorly because it always prioritizes goal-oriented motions. 
In contrast, SC-GOC resolves these instances faster by favoring sideway motions, which break livelock and improve solution quality (\cref{fig:ablation_clustering}). 
For these instances, db-LaCAM with vanilla selection of elements per cluster fails to find a solution within the timelimit. 
This is because the motion set contains many similar motions as described in~\cref{subsec:motion_clustering}; since db-LaCAM performs an exhaustive search, it explores all possible motion combinations, resulting in longer runtimes.

\section{Conclusion}
\label{sec:conclusion}
We introduce db-LaCAM, a planner that combines the scalability of discrete MAPF techniques with the dynamic feasibility of kinodynamic planning. 
By allowing bounded discontinuities between motion primitives, db-LaCAM enables efficient and flexible trajectory generation for multi-robot systems of arbitrary dynamics.
Built upon the coordination mechanism of db-PIBT, it combines local dynamic feasibility with long-horizon search, achieving order-of-magnitude speedups over existing kinodynamic planners while maintaining comparable solution quality.
The successful deployment of flying and car-with-trailer robots highlights the approach’s potential for real-world multi-robot coordination.

\paragraph{Limitations and Future Work}
Our planner db-LaCAM is an informed search-based planner guided by a heuristic $h$.
If the heuristic is imprecise, the planner can get stuck in local minima.
While using the hierarchical EST to estimate $h$ works well for the dynamics considered here, it struggles with more complex systems such as cars with trailers—for example, computing $h$ for the \emph{swap} problem with four car-with-trailer robots takes $6.40~\si{s}$ of a total $6.47~\si{s}$ runtime, as HEST must run for nearly all expanded nodes. 
Moreover, the current approach is centralized and not designed for real-time execution. 
Future work will explore learning-based heuristics for better efficiency, as in MAPF~\cite{jain2025lagat} and enabling real-time, decentralized execution.

\section*{Acknowledgments}
This research was partially funded by the Deutsche Forschungsgemeinschaft (DFG, German Research Foundation) - 448549715, JST ACT-X (JPMJAX22A1), and JST PRESTO (JPMJPR2513).

\bibliography{sty/ref-macro,ref}
\appendix
\onecolumn
\setcounter{secnumdepth}{2}
\section*{Appendix}

\section {\leftline{Motion Primitives}}
\begin{figure}[H]
\centering
    \includegraphics[width=0.5\textwidth]{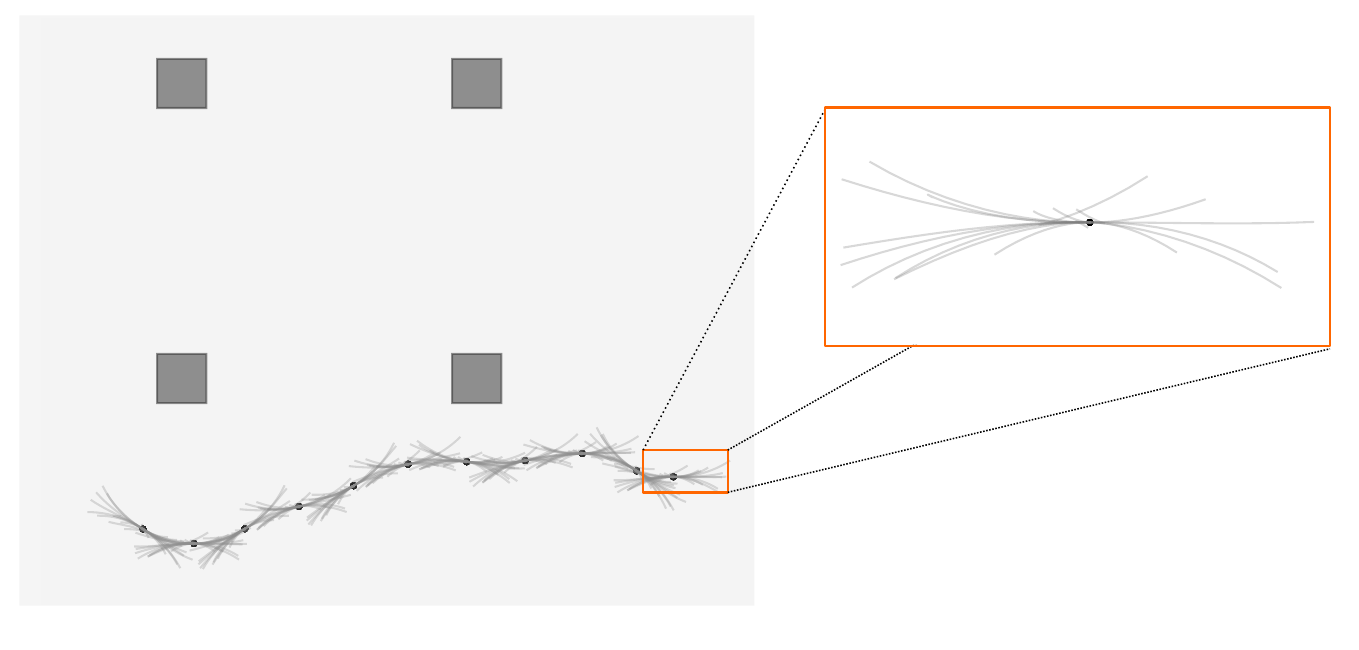}
    \caption{Visualization of motion primitives for unicycle robot dynamics.}
    \label{fig:motion_primitive_visualization}
\end{figure}

We compute motion primitives individually for each robot according to its specific dynamics. 
Unlike lattice-based motions, our motion primitives can take arbitrary shapes. 
 ~\cref{fig:motion_primitive_visualization} shows an example of rolled-out motion primitives for a unicycle robot. 
In this experiment, 200 motion primitives are provided to the planner, but at each timestep, it considers only those applicable to the robot’s current state (black dot).
 Over the course of the test, the planner expands a total of 369 motion primitives.

As demonstrated with the single-robot planner, iDb-A*~\cite{ortiz2024}, motion primitives with discontinuity
search can generalize to quadrotor dynamics with a 13-dimensional state space.
However, as the number of robots increases, the search space grows exponentially
(as in any search-based planning), requiring a more careful selection of motion primitives.

\section{\leftline{Remarks for Thrm. 1}}

LaCAM is a complete algorithm for MAPF, i.e., if a solution exists, it will find it~\cite{okumura2023lacam}. 
LaCAM relies on the argument that, for completeness, the search explores all states that could contain a solution; no potential solution paths are pruned.
For kinodynamic motion planning problems, continuous time and space render full enumeration proofs infeasible, as the set of possible states is infinite.
Instead, we consider resolution-completeness (RC; if a solution exists at the chosen resolution, the algorithm is guaranteed to find it) and probabilistic-completeness (PC; the probability of finding a solution if one exists is 1).

\cref{tab:properties_table} summarizes properties for db-LaCAM based on the motion clustering technique and set of motion primitives.

\begin{remark}  
    By relaxing clustering conditions, such as disabling clustering in~\cref{alg:dblacam}, db-LaCAM can achieve RC with respect to motion primitives.
\end{remark}

\begin{remark}
    With an iterative run of db-LaCAM, where with each iteration more motion primitives are added to the set $\sM$, db-LaCAM can achieve PC.
    A growing set of motion primitives $\sM$ results in a larger search graph, enabling exploration of all reachable states within the closed search space.
\end{remark}

\begin{table}[h!]
\centering
\begin{tabular}{ccc}
\toprule
Motion Clustering Method & Fixed Motion Set & Incremental Motion Set \\
\midrule
No Cluster  & RC & PC \\
Deterministic & Incomplete & Incomplete \\
Probabilistic & PRC & PC \\
\bottomrule
\end{tabular}
\caption{Summary of properties of db-LaCAM.}
\label{tab:properties_table}
\end{table}

\section{\leftline{Proof of Proposition 1}}
\begin{proofsketch}
    For each fixed-horizon planning, the procedure db-PIBT is called $N$ times, once for each robot.
    This is because, db-PIBT($i, \ ..$) is called if and only if the robot $i$ has no reserved motion for the next horizon (Line~\ref{pibtline:call-func}).
    The collision checking between considered motion and high-priority robot motions is $\mathcal{O}(N)$ in the worst case (Line~\ref{pibtline:collision_planned}), resulting in $\mathcal{O}(MN)$ for the loop (Line~\ref{pibtline:for_loop}-Line~\ref{pibtline:return_valid}).
    The collision checking against potential motions of low-priority robots can be $\mathcal{O}(MNM_{max})$ (Line~\ref{pibtline:collision_unplanned}), where $M_{max}$ is the maximum number of motions among all robots.
    Combining these two operations gives $\mathcal{O}(NM^2)$, assuming all robots have the same number of motions.
\end{proofsketch}

\section{\leftline{Procedure \texttt{Set\_Constraint\_Tree} in~\cref{alg:dblacam}}}
Unlike LaCAM, where constraints correspond to occupied vertices, db-LaCAM defines constraints as valid motions.
\cref{alg:update_constraint_tree} illustrates how the constraint tree of a high-level node $Q$ is updated given a set of valid motions $\boldsymbol{M}$.
Constraints are assigned to robots according to their priority order (Line~\ref{uctline:depth_check}). 
Once a robot is identified, the algorithm loops over its valid motions (Line~\ref{uctline:loop_motions}), creating a new constraint tree $C_{new}$ for each motion (Line~\ref{uctline:create_tree}).
Each newly created constraint tree is then added to the constraint trees of $Q$.
If this high-level node $Q$ is picked in further iterations of db-LaCAM (\cref{alg:dblacam}, Line~\ref{line:pop_node}), these constraints are respected during motion generation with db-PIBT (\cref{alg:dblacam}, Line~\ref{line:motion_generator}). 

\begin{algorithm}[ht]
\caption{Set\_Constraint\_Tree}
\label{alg:update_constraint_tree}
\begin{algorithmic}[1]
\Input{processed motion set $\boldsymbol{\hat{M}}$, high-level node $Q$, constraint tree $C$}
\Output{updated constraint tree for $Q$}
\State{$N = |Q.state|$} \Comment{Number of robots}
\If{$\texttt{depth}(C) \leq N$} 
      \State{$Q.order = \texttt{Get\_Order}(Q)$} \Comment{assign robot priority order}
      \State $i \leftarrow Q.order[\texttt{depth}(C)]$ \label{uctline:depth_check}
      \label{algo:planner:selection}
      \State{$\hat{\sM}^i = \boldsymbol{\hat{M}[i]}$} \Comment{get valid motions}
      \For{$m \in \hat{\sM}^i$} \Comment{each valid motion as constraint} \label{uctline:loop_motions}
      \State{$\vx = m.\vx_f$}
      \State $C_{new} \leftarrow \langle~parent: C, who: i, where: \vx, motion: m~\rangle$ \label{uctline:create_tree}
      \State $Q.tree.{push}(C_{new})$
      \EndFor
      \EndIf

\end{algorithmic}
\end{algorithm}

\section{\leftline{Details of Hierarchical EST from \cref{subsec:heuristics}}}

\begin{figure}[H]
\centering
    \includegraphics[width=0.5\textwidth]{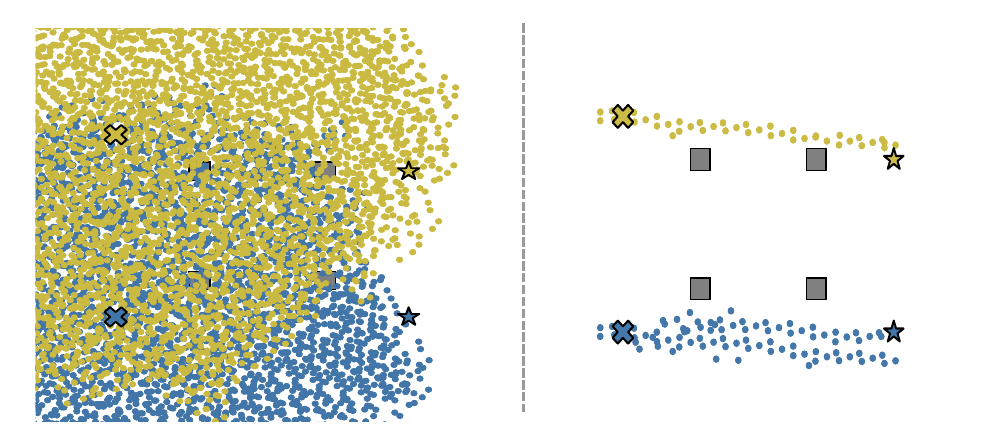}
    \caption{Visualization of state expansions for heuristic $h$ computation: reverse search from goals to starts using db-A*, and forward search with EST. Crosses and stars mark each robot’s start and goal; explored states are shown as colored dots per robot.}
    \label{fig:ablation_node_expansion}
\end{figure}

As can be seen in~\cref{fig:ablation_node_expansion} (left), reverse search with db-A* can lead to exhaustive exploration of states that are irrelevant for finding a solution.
In contrast, \emph{HEST} explores states that can be used to find a solution from the start to the goal state efficiently.

\section{\leftline{Hyperparameter Values for Benchmarking (\cref{sec:validation})}}
The number of motion primitives used for the search is identical for all problem instances and equal to $300$.
The value of the discontinuity bound $\delta$ is 0.5 for all instances.
The value of the distance threshold $\Delta$ for the lookup table is $1.0$ for all dynamics.
The range value for the GOC clustering method is $0.05$ for unicycle dynamics, $1.0$ for 3D double integrator, and the car with trailer.
The distance threshold $\tau$ for the SC-GOC clustering method is set to $1.0$ for all dynamics.

\section{\leftline{Evaluating Performance of db-PIBT}}
While db-PIBT offers high efficiency, its incompleteness makes it difficult to resolve livelock situations, which motivated the development of db-LaCAM.
We compare the success rate of db-PIBT against db-LaCAM with problem instances from~\cref{fig:main}.
\begin{figure}[H]
\centering
    \includegraphics[width=0.75\textwidth]{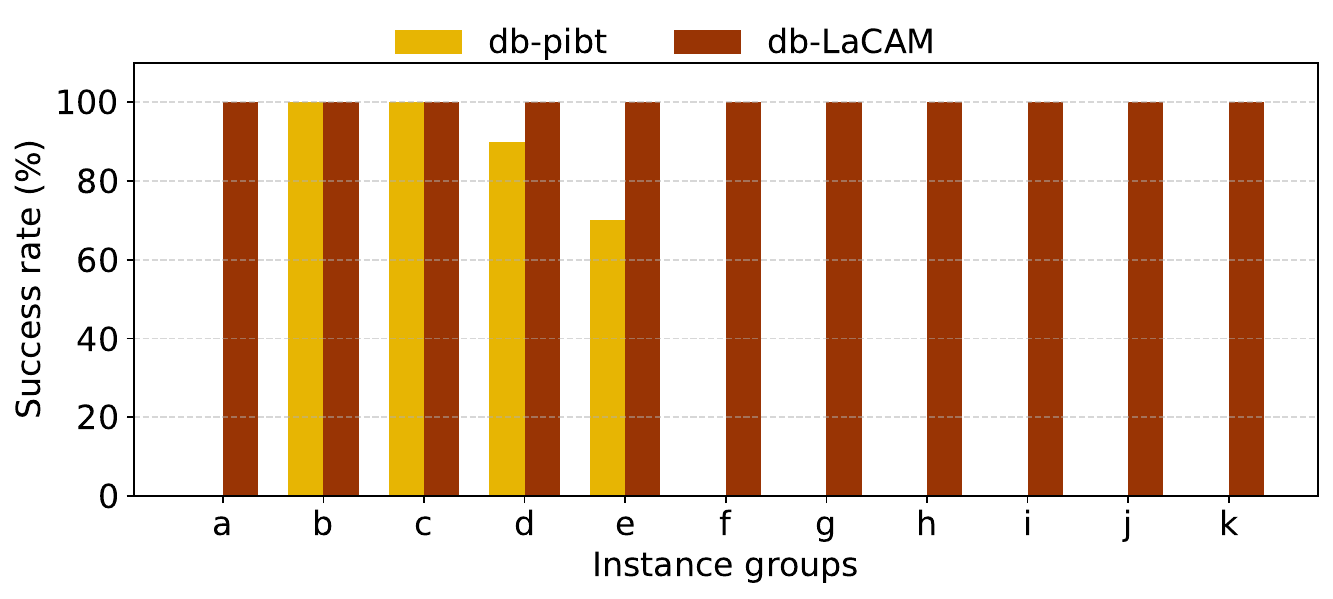}
    \caption{Success rate comparison between db-PIBT and db-LaCAM across 10 instance groups (a–k) shown in~\cref{fig:main}. Each bar represents the success rate over 10 trials per group.}
    \label{fig:success_rate}
\end{figure}
\end{document}